\pdfoutput=1

\documentclass[11pt]{article}

\usepackage[final]{acl}

\usepackage{times}
\usepackage{latexsym}
\usepackage{amsmath}
\usepackage{amssymb}
\usepackage{mathtools}
\usepackage{amsthm}
\usepackage{booktabs}
\usepackage{multirow}
\usepackage{graphicx}

\theoremstyle{plain}

\theoremstyle{definition}

\theoremstyle{remark}

\usepackage[T1]{fontenc}

\usepackage[utf8]{inputenc}

\usepackage{microtype}

\usepackage{inconsolata}

\usepackage{graphicx}
\usepackage{algorithm}
\usepackage{algorithmic}

\usepackage{multirow} 
\usepackage{enumitem}
\usepackage{subcaption}
\usepackage{pgf-pie}
\usepackage{graphicx}
\usepackage{caption}
\usepackage{pifont}
\usepackage[table,xcdraw]{xcolor}
\usepackage[normalem]{ulem}
\useunder{\uline}{\ul}{}
\usepackage[textsize=tiny]{todonotes}
\usepackage[skins,breakable,most]{tcolorbox}

\definecolor{darkgreen}{rgb}{0.0, 0.3, 0.0}

\usepackage{xspace}
\newcommand{\muddymaze}{\texttt{MuddyMaze}\xspace}

\title{Dialogue is Better Than Monologue: Instructing Medical LLMs via Strategical Conversation}

\author{
Zijie Liu$^{1}$, Xinyu Zhao$^{1}$, Jie Peng$^{2}$, Jinhao Duan$^{1}$, Zhuangdi Zhu$^{3}$ \\
\textbf{Qingyu Chen}$^{4}$, \textbf{Kaidi Xu}$^{5}$, \textbf{Xia Hu}$^{6}$, \textbf{Tianlong Chen}$^{1}$
\\[0.4em]
$^{1}$UNC at Chapel Hill \quad
$^{2}$University of Science and Technology of China \quad
$^{3}$George Mason University \\
$^{4}$Yale University \quad
$^{5}$City University of Hong Kong \quad
$^{6}$Rice University
\\[0.3em]
Correspondence: \texttt{tianlong@cs.unc.edu}
}

\begin{document}
\maketitle
\begin{abstract}
In real clinical practice, clinicians must sift through noisy and often conflicting information, progressively gathering and sequencing evidence before reaching conclusions. However, existing tuning methods for medical AI models are typically \texttt{monologue-based} — that is, models are fine-tuned on static question-answering (QA) tasks or medical articles, which fail to reflect the interactive and iterative nature of clinical reasoning.
To bridge this gap, we introduce \muddymaze, a benchmark designed to expose the limitations of current \texttt{monologue-based} tuning, and construct a large \texttt{dialogue dataset} of 22.2k doctor–patient interactions that capture stepwise diagnostic reasoning validated by medical experts. Building on those, we propose \texttt{dialogue-tuning}, a new fine-tuning paradigm that captures the internal reasoning dynamics unfolding across interactions.

To assess the effectiveness of our approach, we evaluated \textit{dialogue-tuned} models on \muddymaze, where they outperform \textit{monologue-tuned} baselines (e.g., MedQA) by +16.1\% in one-round and +4.1\% in multi-round evidence ranking, while maintaining or even improving accuracy on standard medical QA benchmarks (e.g., PubMedQA). These results indicate that \texttt{dialogue-tuning} not only enhances reasoning robustness and evidence integration but also preserves the factual precision of traditional QA performance.
\end{abstract}

\section{Introduction} \label{sec:intro}
Large language models (LLMs) have achieved remarkable progress in the medical domain, contributing to applications such as disease analysis, diagnostic support, and clinical decision assistance~\cite{singhal2023large,li2023chatdoctor,li2023doctorgpt,chen2023meditron,peng2023study,kwon2024large}. Despite these advances, the evaluation of medical LLMs remains an open challenge. Most existing benchmarks are derived from medical licensing exams or research articles and are framed as multiple-choice QA or long-form reasoning tasks ~\cite{jin2021disease,pal2022medmcqa}. These benchmarks have provided valuable insights into model knowledge and improved QA performance, but they remain highly structured and artificially clean, diverging from the complexities of real-world diagnostic practice~\cite{chen2024benchmarking,yao2024medqa}. 

\textbf{Limitations of current benchmarks.}
Current medical QA datasets neglect two essential properties of clinical reasoning.
(i) \textit{Stepwise and iterative reasoning.} In real practice, clinicians rarely have all information available upfront. Instead, they iteratively refine hypotheses by actively gathering information, asking targeted questions, and updating decisions as new evidence arrives. By contrast, medical QA benchmarks present complete information in a single turn, testing only the final interpretation of facts while ignoring the process of strategically acquiring and organizing evidence under uncertainty.
(ii) \textit{Noise and uncertainty.} Clinical data are often incomplete, ambiguous, or even conflicting—for instance, vague symptoms or false-positive test results. Physicians must weigh and prioritize evidence while resisting misleading signals. Current benchmarks, however, use well-structured case descriptions and offer no mechanism for assessing how models handle distractors, uncertainty, or noisy clinical contexts.
\begin{figure*}[t]
    \centering
    \includegraphics[scale=0.43]{figure/pipeline.pdf}
    \vspace{-6mm}
    \caption{\small (Top) \muddymaze reframes static medical QA as evidence ranking, evaluating reasoning via one-round and multi-round modes.
(Bottom) Dialogue-Tuning converts QA and article data into doctor–patient dialogues and fine-tunes models only on physician turns, encouraging stepwise clinical reasoning.}
    \label{fig:benchmark}
    \vspace{-3mm}
\end{figure*}

\textbf{Our Benchmark: \muddymaze.}
We address these gaps with \muddymaze, an answer-conditioned benchmark that reframes medical QA as evidence ranking under noise. 
Each instance comprises background context, a diagnostic QA pair, and an evidence pool containing both relevant and distracting findings. 
Moreover, \muddymaze evaluates reasoning through one-round and multi-round evidence ranking, simulating stepwise information gathering under varying noise and difficulty levels (see Section \ref{sec:benchmark} for details). 
This design enables interpretable measurement of how models reason under uncertainty rather than in perfectly structured settings.

\textbf{Dialogue-based training.} 
While \muddymaze evaluates reasoning, training remains limited if models continue to learn only from static QA objectives that provide final-answer supervision. 
To approximate clinicians’ stepwise reasoning, we introduce a dialogue-based fine-tuning paradigm that reformulates medical QA and article cases into doctor–patient interactions. 
During training, the model generates doctor responses conditioned on evolving conversational context, thereby learning to reason iteratively rather than memorizing direct QA mappings. 
Dialogue-based training encourages models to organize and prioritize evidence more effectively in noisy scenarios, leading to consistent improvements on \muddymaze and stable performance on QA benchmarks.

\noindent $\bullet$ We introduce \muddymaze, a benchmark that transforms static QA into stepwise evidence ranking under noise, with both one-round and multi-round evaluation and tiered difficulty levels.

\noindent $\bullet$ We curate a large-scale dialogue dataset simulating doctor–patient interactions, enabling models to learn iterative reasoning strategies.

\noindent $\bullet$ We propose a dialogue-based fine-tuning framework that enhances evidence-ranking performance as a proxy for reasoning quality, while maintaining strong results on conventional QA benchmarks.

\section{Related Work}
\textbf{Medical Large Language Models.} 
LLMs have shown rapid progress in the medical domain \cite{singhal2023large,chen2023meditron,wu2023pmcllamabuildingopensourcelanguage,zhang2024alpacareinstructiontunedlargelanguagemodels,gema2024parameterefficientfinetuningllamaclinical,han2023medalpacaopensourcecollection,xie2024llamafoundationlargelanguage}, achieving state-of-the-art results on medical QA and summarization tasks. 
These models are typically fine-tuned on medical exams, scientific literature, clinical guidelines, and EHR notes to adapt to real-world clinical text understanding. 
In parallel, bio-focused language models have been developed for biomedical research and healthcare applications \cite{luo2022biogpt,bannur2023learning}. 
For example, BioBERT \cite{lee2020biobert} and PubMedBERT \cite{gu2021domain}, both pre-trained on PubMed, serve as foundational models for tasks such as named entity recognition and relation extraction.

\textbf{Conversation Datasets in the Medical Field.}
Early medical benchmarks mainly assessed factual knowledge through static QA formats such as multiple-choice or extractive tasks, exemplified by MedQA \cite{yao2024medqa}, MedMCQA \cite{pal2022medmcqa}, and PubMedQA \cite{jin2019pubmedqa}.  
Recent work has shifted toward dialogue-based settings that reflect real doctor–patient communication, including MedDialog \cite{zeng2020meddialog} and ReMeDi \cite{yan2022remedi}.  
Several domain-specific datasets further extend this paradigm, covering mental health counseling \cite{chen2024convcounsel}, pediatrics \cite{zhang2024pediabench}, and COVID-19 consultations \cite{ju2020coviddialog}.  
More recently, studies such as MediQ \cite{li2024mediqquestionaskingllmsbenchmark} and UoT \cite{hu2024uncertaintythoughtsuncertaintyawareplanning} explored interactive information-seeking dialogues, where models actively query for missing clinical details—highlighting a growing focus on reasoning-oriented medical conversations.

\vspace{-1mm}
\section{Methodology} \label{sec:method}
\vspace{-1mm}
This section presents our methodology for enhancing clinical reasoning in medical LLMs through \texttt{dialogue-tuning}. 
We reformulate standard medical QA and article-based data into synthetic doctor–patient dialogues, where the model learns to generate physician responses conditioned on evolving conversational context. 
By fine-tuning on the physician’s turns, \texttt{dialogue-tuning} encourages stepwise reasoning behaviors that better reflect real clinical decision-making processes.

\subsection{From Monologue to Dialogue Tuning} \label{sec:method_1}

\textbf{Monologue Tuning.} 
Most medical training datasets are \texttt{monologue-style}, such as multi-choice QA or article-based tasks, where all relevant information is given upfront and the model is trained to output a final answer. While effective for knowledge recall, this format oversimplifies clinical reasoning, which in reality is iterative, uncertain, and evidence-driven.

\textbf{Dialogue Tuning.} 
Dialogue-style supervision better mirrors real diagnostic workflows. In doctor–patient interactions, information is revealed progressively rather than all at once. At each step, the clinician integrates known facts, asks targeted questions, and updates diagnostic hypotheses. 
To approximate this process, we reformulate training data into multi-turn dialogues where the model generates physician responses conditioned on prior context. Each doctor turn requires the model to (i) identify relevant findings from the patient’s utterances and (ii) determine the next appropriate inquiry or interpretation. This setup provides structured supervision for iterative evidence integration and hypothesis refinement—key aspects of clinical reasoning absent in standard \texttt{monologue-style} training. A detailed comparison between \texttt{monologue-} and \texttt{dialogue-tuning} is provided in Figure~\ref{fig:benchmark}.
\begin{figure}[]
    \centering
    \includegraphics[scale=0.15]{figure/datainfo.pdf} 
    \vspace{-6mm}
    \caption{\small The left pie chart represents the ratio of difficulty levels in our benchmark. While the right pie chart represents the proportion of multiple-choice question-answering sets and articles used during the tuning stage, the dialogues generated from these sources are equal in quantity to them.}
    \label{fig:datainfo}
    \vspace{-4mm}
\end{figure}

\subsection{Dialogue Tuning Objective} \label{sec:method_2}
To better align training with clinical reasoning, we convert existing datasets into synthetic doctor–patient dialogues (see \autoref{fig:datainfo}). Two sources are used: 10.2k QA pairs from the MedQA ~\cite{jin2021disease}  training set and 12k articles from PubMed. 
\ding{182} \textit{Multi-Choice to Dialogue}: Each QA item is reformulated into a dialogue where the patient introduces demographics, symptoms, or test results, and the physician probes further and provides interpretations. 
\ding{183} \textit{Article to Dialogue}: Article Cases are similarly decomposed into exchanges: patients describe their history and findings in accessible language, while doctors interpret, ask follow-up questions, and reason about possible diagnoses. 
In both cases, technical details are preserved but revealed turn by turn, allowing the dialogue to unfold progressively and mirror real diagnostic encounters where physicians refine hypotheses step by step.

\textbf{Implementation.} We employ Gemini-2.5 Flash ~\cite{li2024mediqquestionaskingllmsbenchmark} to perform the conversion, using carefully designed prompts to ensure faithfulness to the source material (\autoref{sec:appA}). Dialogue length and number of turns vary with sample complexity. To assess quality, we conduct human evaluation on a stratified sample, confirming that essential information is preserved (\autoref{sec:human_eval}).  

Our \texttt{dialogue-tuning} approach trains the LLM to generate physician responses conditioned on the previous conversational context. This differs fundamentally from conventional next-token prediction by operating at the level of complete dialogue acts rather than individual tokens.

\textbf{Training Setup.} In \texttt{dialogue-tuning}, the model plays the role of the physician. Given a dialogue prefix 
\vspace{-3mm}
\[
H_t = \{u_1, d_1, \ldots, u_{t-1}, u_t\},
\]

where $u_i$ and $d_i$ denote patient and physician's turns, the model predicts the next physician utterance $d_t$. The objective is standard token-level cross-entropy:  
\vspace{-4mm}
\begin{equation}
\mathcal{L} = -\sum_{t \in \mathcal{T}_D} \sum_{i=1}^{|u_t|} \log P(w_i^t | w_{<i}^t, H_t),
\end{equation}

where $\mathcal{T}_D$ indexes physician turns. The loss is computed only on physician responses, conditioning on the full conversational context. Although the optimization objective remains unchanged, \texttt{dialogue-tuning} differs conceptually from \texttt{monologue-tuning} by training models to generate reasoning steps rather than final answers.

\section{\muddymaze: Benchmark for Evidence Ranking} \label{sec:benchmark}
To evaluate reasoning under realistic clinical uncertainty, we introduce \muddymaze, a benchmark that reformulates QA into structured evidence-ranking tasks (\autoref{fig:benchmark}). Each instance comprises: (i) \textit{background information} (e.g., patient history and symptoms), (ii) a \textit{diagnostic QA pair} specifying the target problem, and (iii) an \textit{evidence pool} containing gold supporting sentences mixed with clinically plausible distractors.  
The model must output an \textit{ordered evidence sequence} that connects the background to the answer while filtering out irrelevant information. This formulation transforms static QA into a reasoning-by-ranking problem, directly testing whether models can identify, prioritize, and organize evidence in noisy clinical contexts.  

\subsection{Dataset Sources}
\muddymaze\ is constructed by reformulating multiple publicly available medical QA benchmarks into an evidence-ranking format. For each instance, we decompose the original QA item into background, QA-pair and evidence sentences. Moreover, distractors are introduced by sampling clinically plausible but irrelevant content to ensure robustness against noise. Specifically:  
\ding{182} \textbf{MedQA  Test Set (USMLE Step 1–3) ~\cite{jin2021disease}}: Provides foundational and advanced clinical questions.   
\ding{183} \textbf{MedBullets Step 2/3 ~\cite{chen2025benchmarkinglargelanguagemodels}}: Offers high-quality preparation questions that require integrating multiple evidence pieces. These items enrich the benchmark’s advanced-level reasoning scenarios.  
\ding{184} \textbf{JAMA Challenge ~\cite{chen2025benchmarkinglargelanguagemodels}}: Contains complex, real-world case vignettes from the \textit{Journal of the American Medical Association}, representing the most challenging and ambiguous diagnostic reasoning tasks.

\subsection{Difficulty Levels Aligned with USMLE}
To mirror the progressive complexity of medical training, \muddymaze\ is organized into three tiers aligned with USMLE stages: 

\noindent $\bullet$ \textbf{Basic Level.} Derived from MedQA Step 1 items focusing on single-fact biomedical reasoning.  

\noindent $\bullet$ \textbf{Advanced Level.} Based on Step 2/3 and MedBullets questions requiring multi-evidence synthesis across history, labs, and guidelines. 

\noindent $\bullet$ \textbf{Challenge Level.} Formed from JAMA Challenge cases with ambiguous findings and subtle distractors, requiring prioritization under uncertainty.  

\subsection{Noise Levels Reflecting Real-World Uncertainty}
In clinical practice, not all available information is reliable or relevant. To capture this, \muddymaze introduces noise levels by injecting distractor sentences into the evidence pool.  

\noindent $\bullet$ \textbf{Level 0 (Clean).} No distractors are added; all evidence is directly relevant to the diagnostic task.  

\noindent $\bullet$ \textbf{Level $n > 0$ (Noisy).} A specified number of distractors are sampled from unrelated clinical contexts and inserted into the pool alongside the gold supporting evidence. Distractors may appear clinically plausible but do not contribute to the correct reasoning chain.  

This design assesses a model’s ability to sustain reasoning performance under uncertainty. By introducing controlled noise, the benchmark evaluates whether models can accurately identify and prioritize informative evidence while disregarding misleading or irrelevant inputs..

\begin{algorithm}[tb]
   \caption{Multi Round \muddymaze}
   \label{alg:mr}
\begin{algorithmic}
   \STATE {\bfseries Input:} Background Information $BI$, Question $Q$, Answer $A$, Evidence Sentences $E = \{e_1, e_2, \ldots, e_n\}$, Total Attempts $T$
   \STATE Initialize $i \gets \emptyset$ \COMMENT{No sentence selected yet}
   \FOR{$t=1$ {\bfseries to} $T$}
   \IF{$t = 1$}
   \STATE Display current $BI$, $Q$, $A$, and $E$
   \STATE Prompt model to select a single sentence index $i_t \in \{1, 2, \ldots, n\}$
   \ELSE
   \STATE Update $BI \gets BI \cup e_{i_{t-1}}$ \COMMENT{Add previous sentence to Background Information}
   \STATE Display updated $BI$, $Q$, $A$, and $E$
   \STATE Prompt model to select a single sentence index $i_t \in \{1, 2, \ldots, n\}$
   \ENDIF
   \IF{$t = T$}
   \STATE {\bfseries break} \COMMENT{Final attempt reached}
   \ENDIF
   \ENDFOR
   \STATE {\bfseries Return} $i_T$ \COMMENT{Final selected sentence index}
\end{algorithmic}
\end{algorithm}

\begin{table*}[]
\vspace{-3mm}
\resizebox{\textwidth}{!}{%
\begin{tabular}{@{}ccccccccccc@{}}
\toprule
\multicolumn{11}{c}{Basic (Multi-Hop Acc)} \\ \midrule
Model & \multicolumn{1}{c|}{Noise Level} & Raw & Multi-Choice & \multicolumn{1}{c|}{Dialogue(MC)} & Raw & Article & \multicolumn{1}{c|}{Dialogue(Article)} & Raw & Baseline & Combined Dialogue \\ \midrule
 & \multicolumn{1}{c|}{0} & 0.2707 & 0.1473 & \multicolumn{1}{c|}{\textbf{0.3253}} & 0.2707 & 0.3221 & \multicolumn{1}{c|}{\textbf{0.3317}} & 0.2707 & 0.2464 & \textbf{0.3335} \\
 & \multicolumn{1}{c|}{1} & 0.3526 & 0.1114 & \multicolumn{1}{c|}{\textbf{0.4779}} & 0.3526 & 0.4477 & \multicolumn{1}{c|}{\textbf{0.5224}} & 0.3526 & 0.302 & \textbf{0.4882} \\
\multirow{-3}{*}{Llama-3.2-3B-Instruct} & \multicolumn{1}{c|}{3} & 0.2482 & 0.1232 & \multicolumn{1}{c|}{\textbf{0.2842}} & 0.2482 & 0.3269 & \multicolumn{1}{c|}{\textbf{0.3754}} & 0.2482 & 0.2545 & \textbf{0.3919} \\
 & \multicolumn{1}{c|}{0} & 0.2804 & 0.2101 & \multicolumn{1}{c|}{\textbf{0.3289}} & 0.2804 & \textbf{0.2993} & \multicolumn{1}{c|}{0.271} & 0.2804 & 0.2183 & \textbf{0.3045} \\
 & \multicolumn{1}{c|}{1} & 0.3715 & 0.2198 & \multicolumn{1}{c|}{\textbf{0.4249}} & 0.3715 & 0.3697 & \multicolumn{1}{c|}{\textbf{0.3779}} & 0.3715 & 0.254 & \textbf{0.4007} \\
\multirow{-3}{*}{Qwen2.5-3B-Instruct} & \multicolumn{1}{c|}{3} & 0.3337 & 0.2023 & \multicolumn{1}{c|}{\textbf{0.3404}} & \textbf{0.3337} & 0.3262 & \multicolumn{1}{c|}{0.3259} & 0.3337 & 0.2746 & \textbf{0.343} \\
\rowcolor[HTML]{CFF7F2} 
\multicolumn{2}{c|}{\cellcolor[HTML]{CFF7F2}Average} & 0.3095 & 0.1690 & \multicolumn{1}{c|}{\cellcolor[HTML]{CFF7F2}\textbf{0.3636}} & 0.3095 & 0.3487 & \multicolumn{1}{c|}{\cellcolor[HTML]{CFF7F2}\textbf{0.3674}} & 0.3095 & 0.2583 & \textbf{0.3770} \\ \midrule
\multicolumn{11}{c}{Basic (Single-Wise Acc)} \\ \midrule
Model & \multicolumn{1}{c|}{Noise Level} & Raw & Multi-Choice & \multicolumn{1}{c|}{Dialogue(MC)} & Raw & Article & \multicolumn{1}{c|}{Dialogue(Article)} & Raw & Baseline & Combined Dialogue \\ \midrule
 & \multicolumn{1}{c|}{0} & 0.4104 & 0.1991 & \multicolumn{1}{c|}{\textbf{0.4528}} & 0.4104 & 0.4624 & \multicolumn{1}{c|}{\textbf{0.4635}} & 0.4104 & 0.2397 & \textbf{0.4629} \\
 & \multicolumn{1}{c|}{1} & 0.4171 & 0.15 & \multicolumn{1}{c|}{\textbf{0.5104}} & 0.4171 & 0.4891 & \multicolumn{1}{c|}{\textbf{0.561}} & 0.4171 & 0.2421 & \textbf{0.5044} \\
\multirow{-3}{*}{Llama-3.2-3B-Instruct} & \multicolumn{1}{c|}{3} & 0.2856 & 0.1303 & \multicolumn{1}{c|}{\textbf{0.3096}} & 0.2856 & 0.3389 & \multicolumn{1}{c|}{\textbf{0.3587}} & 0.2856 & 0.1951 & \textbf{0.3552} \\
 & \multicolumn{1}{c|}{0} & 0.3845 & 0.298 & \multicolumn{1}{c|}{\textbf{0.4239}} & 0.3845 & \textbf{0.3836} & \multicolumn{1}{c|}{0.339} & \textbf{0.3845} & 0.2642 & 0.3735 \\
 & \multicolumn{1}{c|}{1} & 0.5434 & 0.4427 & \multicolumn{1}{c|}{\textbf{0.5568}} & 0.5434 & \textbf{0.5364} & \multicolumn{1}{c|}{0.5137} & 0.5434 & 0.4569 & \textbf{0.5347} \\
\multirow{-3}{*}{Qwen2.5-3B-Instruct} & \multicolumn{1}{c|}{3} & \textbf{0.4023} & 0.2731 & \multicolumn{1}{c|}{0.3717} & 0.4023 & \textbf{0.3797} & \multicolumn{1}{c|}{0.3671} & \textbf{0.4023} & 0.3314 & 0.3733 \\
\rowcolor[HTML]{CFF7F2} 
\multicolumn{2}{c|}{\cellcolor[HTML]{CFF7F2}Average} & 0.4072 & 0.2489 & \multicolumn{1}{c|}{\cellcolor[HTML]{CFF7F2}\textbf{0.4375}} & 0.4072 & 0.4317 & \multicolumn{1}{c|}{\cellcolor[HTML]{CFF7F2}\textbf{0.4338}} & 0.4072 & 0.2882 & \textbf{0.4340} \\ \bottomrule
\end{tabular}%
}
\caption{\small Performance of Llama-3.2-3B-Instruct and Qwen2.5-3B-Instruct under three tuning strategies across noise levels, evaluated on Basic tasks in One-Round setting.}
\label{tab:basic}
\vspace{-2mm}
\end{table*}

\begin{table*}[h]
\resizebox{\textwidth}{!}{%
\begin{tabular}{@{}ccccccccccc@{}}
\toprule
\multicolumn{11}{c}{Advance (Multi-Hop Acc)} \\ \midrule
Model & \multicolumn{1}{c|}{Noise Level} & Raw & Multi-Choice & \multicolumn{1}{c|}{Dialogue(MC)} & Raw & Article & \multicolumn{1}{c|}{Dialogue(Article)} & Raw & Baseline & Combined Dialogue \\ \midrule
 & \multicolumn{1}{c|}{0} & 0.1092 & 0.052 & \multicolumn{1}{c|}{\textbf{0.1578}} & 0.1092 & 0.1574 & \multicolumn{1}{c|}{\textbf{0.1588}} & 0.1092 & 0.0867 & \textbf{0.1683} \\
 & \multicolumn{1}{c|}{1} & 0.1815 & 0.0522 & \multicolumn{1}{c|}{\textbf{0.424}} & 0.1815 & 0.3965 & \multicolumn{1}{c|}{\textbf{0.4804}} & 0.1815 & 0.1995 & \textbf{0.4308} \\
\multirow{-3}{*}{Llama-3.2-3B-Instruct} & \multicolumn{1}{c|}{3} & 0.1508 & 0.0487 & \multicolumn{1}{c|}{\textbf{0.2198}} & 0.1508 & 0.2474 & \multicolumn{1}{c|}{\textbf{0.3014}} & 0.1508 & 0.1539 & \textbf{0.2925} \\
 & \multicolumn{1}{c|}{0} & 0.1449 & 0.1374 & \multicolumn{1}{c|}{\textbf{0.1667}} & 0.1449 & 0.154 & \multicolumn{1}{c|}{\textbf{0.1603}} & 0.1449 & 0.1207 & \textbf{0.1668} \\
 & \multicolumn{1}{c|}{1} & 0.339 & 0.2375 & \multicolumn{1}{c|}{\textbf{0.479}} & 0.339 & \textbf{0.3873} & \multicolumn{1}{c|}{0.3765} & 0.339 & 0.2888 & \textbf{0.4496} \\
\multirow{-3}{*}{Qwen2.5-3B-Instruct} & \multicolumn{1}{c|}{3} & \textbf{0.2995} & 0.2108 & \multicolumn{1}{c|}{0.2965} & 0.2995 & 0.2822 & \multicolumn{1}{c|}{\textbf{0.3}} & 0.2995 & 0.2391 & \textbf{0.315} \\
\rowcolor[HTML]{CFF7F2} 
\multicolumn{2}{c|}{\cellcolor[HTML]{CFF7F2}Average} & 0.2042 & 0.1231 & \multicolumn{1}{c|}{\cellcolor[HTML]{CFF7F2}\textbf{0.2906}} & 0.2042 & 0.2708 & \multicolumn{1}{c|}{\cellcolor[HTML]{CFF7F2}\textbf{0.2962}} & 0.2042 & 0.1815 & \textbf{0.3038} \\ \midrule
\multicolumn{11}{c}{Advance (Single-Wise Acc)} \\ \midrule
Model & \multicolumn{1}{c|}{Noise Level} & Raw & Multi-Choice & \multicolumn{1}{c|}{Dialogue(MC)} & Raw & Article & \multicolumn{1}{c|}{Dialogue(Article)} & Raw & Baseline & Combined Dialogue \\ \midrule
 & \multicolumn{1}{c|}{0} & 0.1926 & 0.0823 & \multicolumn{1}{c|}{\textbf{0.2897}} & 0.1926 & \textbf{0.3154} & \multicolumn{1}{c|}{0.2939} & 0.1926 & 0.1274 & \textbf{0.2937} \\
 & \multicolumn{1}{c|}{1} & 0.3178 & 0.1031 & \multicolumn{1}{c|}{\textbf{0.6159}} & 0.3178 & 0.5859 & \multicolumn{1}{c|}{\textbf{0.6676}} & 0.3178 & 0.2319 & \textbf{0.6092} \\
\multirow{-3}{*}{Llama-3.2-3B-Instruct} & \multicolumn{1}{c|}{3} & 0.2604 & 0.0852 & \multicolumn{1}{c|}{\textbf{0.3754}} & 0.2604 & 0.4021 & \multicolumn{1}{c|}{\textbf{0.4405}} & 0.2604 & 0.1837 & \textbf{0.4232} \\
 & \multicolumn{1}{c|}{0} & 0.2557 & 0.218 & \multicolumn{1}{c|}{\textbf{0.2733}} & 0.2557 & \textbf{0.2714} & \multicolumn{1}{c|}{0.2506} & 0.2557 & 0.1973 & \textbf{0.2713} \\
 & \multicolumn{1}{c|}{1} & 0.5579 & 0.561 & \multicolumn{1}{c|}{\textbf{0.6919}} & 0.5579 & \textbf{0.6681} & \multicolumn{1}{c|}{0.6521} & 0.5579 & 0.6469 & \textbf{0.6777} \\
\multirow{-3}{*}{Qwen2.5-3B-Instruct} & \multicolumn{1}{c|}{3} & 0.4131 & 0.3552 & \multicolumn{1}{c|}{\textbf{0.4368}} & 0.4131 & 0.4309 & \multicolumn{1}{c|}{\textbf{0.4363}} & 0.4131 & 0.4332 & \textbf{0.4511} \\
\rowcolor[HTML]{CFF7F2} 
\multicolumn{2}{c|}{\cellcolor[HTML]{CFF7F2}Average} & 0.3329 & 0.2341 & \multicolumn{1}{c|}{\cellcolor[HTML]{CFF7F2}\textbf{0.4472}} & 0.3329 & 0.4456 & \multicolumn{1}{c|}{\cellcolor[HTML]{CFF7F2}\textbf{0.4568}} & 0.3329 & 0.3034 & \textbf{0.4544} \\ \bottomrule
\end{tabular}%
}
\caption{\small Performance of Llama-3.2-3B-Instruct and Qwen2.5-3B-Instruct under three tuning strategies across noise levels, evaluated on Adavance tasks in One-Round setting.}
\vspace{-5mm}
\label{tab:advance}

\end{table*}

\subsection{Task Settings}
Two complementary settings probe distinct reasoning behaviors:  

\noindent $\bullet$ \textbf{One-Round Evidence Ranking.} The model receives background, QA pair, and the full evidence pool, and must output a complete ordered evidence chain in one step. This setting tests global reasoning organization without intermediate feedback.  

\noindent $\bullet$ \textbf{Multi-Round Evidence Ranking.} At each round $t$, the model selects one evidence sentence, which is appended to the background before the next iteration. The process continues for $T$ rounds, yielding a progressive evidence sequence $\{i_1, i_2, \ldots, i_T\}$. This iterative formulation evaluates stepwise evidence acquisition and integration.  

Together, the two settings assess both holistic reasoning and incremental evidence gathering.  

\subsection{Evaluation Metrics}

\muddymaze\ evaluates model reasoning from both global and local perspectives through two complementary metrics. Each instance consists of background information, a diagnostic QA pair, and an evidence pool from which the model must output an ordered chain of supporting sentences.  

\noindent $\bullet$ \textbf{Multi-Hop Accuracy.}  
This metric measures whether the model exactly reconstructs the annotated reasoning chain, considering both content and position. For a chain of length $N$, let $e_i$ denote the $i$-th gold evidence sentence with position $p_i$. A prediction at position $i$ is correct only if the selected sentence $\hat{e}_i = e_i$ and its assigned position $\hat{p}_i = p_i$. The score is defined as:  

\begin{equation*}
\text{Multi-Hop Accuracy} = \frac{\sum_{i=1}^{N} \mathbb{I}(e_i = \hat{e}_i \wedge p_i = \hat{p}_i)}{N}
\end{equation*}

where $\mathbb{I}$ is the indicator function. This metric provides a strict assessment of whether the model fully recovers the global reasoning chain.  

\noindent $\bullet$ \textbf{Single-Wise Accuracy.}  
This metric evaluates whether the model preserves the adjacency relations between consecutive evidence sentences. A ground-truth pair $(e_i, e_{i+1})$ is considered correct if the predicted sequence contains $(\hat{e}_i, \hat{e}_{i+1}) = (e_i, e_{i+1})$ or its reversed order $(\hat{e}_i, \hat{e}_{i+1}) = (e_{i+1}, e_i)$. Formally:  

{\small
\begin{equation*}
\begin{array}{c}
\text{Single-Wise} \\
\text{Accuracy}
\end{array}
=
\frac{1}{N-1} \sum_{i=1}^{N-1} \mathbb{I}
\begin{pmatrix}
(e_i, e_{i+1}) = (\hat{e}_i, \hat{e}_{i+1}) \lor \\
(e_i, e_{i+1}) = (\hat{e}_{i+1}, \hat{e}_i),
\end{pmatrix}
\end{equation*}
}

This metric reflects the model’s ability to maintain coherent transitions between reasoning steps even when the full order is imperfect, thus highlighting local consistency under uncertainty.

Together, Multi-Hop and Single-Wise Accuracy form a comprehensive evaluation of reasoning quality.  
The former emphasizes exact recovery of the global reasoning structure, while the latter captures the preservation of clinically meaningful intermediate relations, offering a balanced view of both holistic and incremental reasoning performance. 

\section{Experiment}
\vspace{-2mm}
\label{sec:experiment}
This section evaluates the effectiveness of our proposed \texttt{dialogue-tuning} strategy through three controlled comparison settings.

\subsection{Experiments Setup}
\vspace{-2mm}
We conduct experiments to evaluate how dialogue-based training affects reasoning performance across different types of medical data. (i) \textbf{Multiple-Choice Setting.}  
Models are fine-tuned on standard multiple-choice QA data (\textit{Multi-Choice}) and its dialogue-converted variant (\textit{Dialogue(MC)}), compared against the zero-shot \textit{Raw} model.  
This setup tests whether dialogue-style training helps models move beyond pattern recognition toward evidence-based reasoning within structured QA formats.  
(ii) \textbf{Article-Based Setting.}  
Using medical article data, we compare models fine-tuned directly on raw article text (\textit{Article}) versus their dialogue-transformed versions (\textit{Dialogue(Article)}).  
This evaluates whether conversational reformulation enhances reasoning when training on narrative, unstructured medical text.  
(iii) \textbf{Combined Setting.}  
We further integrate both modalities to test generalization.  
A \textit{Baseline} model is trained on the mix of Multi-Choice QA and Article data, while the \textit{Combined Dialogue} model uses their dialogue-converted counterparts.  
This setting examines whether \texttt{dialogue-tuning} benefits scale consistently across heterogeneous data sources.  

All models are evaluated on the proposed \muddymaze benchmark under both \textit{One-Round} and \textit{Multi-Round} reasoning modes. 
Each mode is tested across two noise levels (1, 3) and three difficulty tiers (Basic, Advanced and Challenge), allowing us to systematically assess reasoning robustness under varying uncertainty. 
We employ two backbone models—Llama-3.2-3B-Instruct (LLaMA) ~\cite{grattafiori2024llama3herdmodels} and Qwen2.5-3B-Instruct (Qwen)~\cite{qwen2025qwen25technicalreport}—and report two complementary metrics: \textit{Multi-Hop Accuracy}, which measures global reconstruction of the reasoning chain, and \textit{Single-Wise Accuracy}, which captures local coherence between consecutive evidence steps.

\begin{figure}[h!]
    \centering
    \includegraphics[scale=0.24]{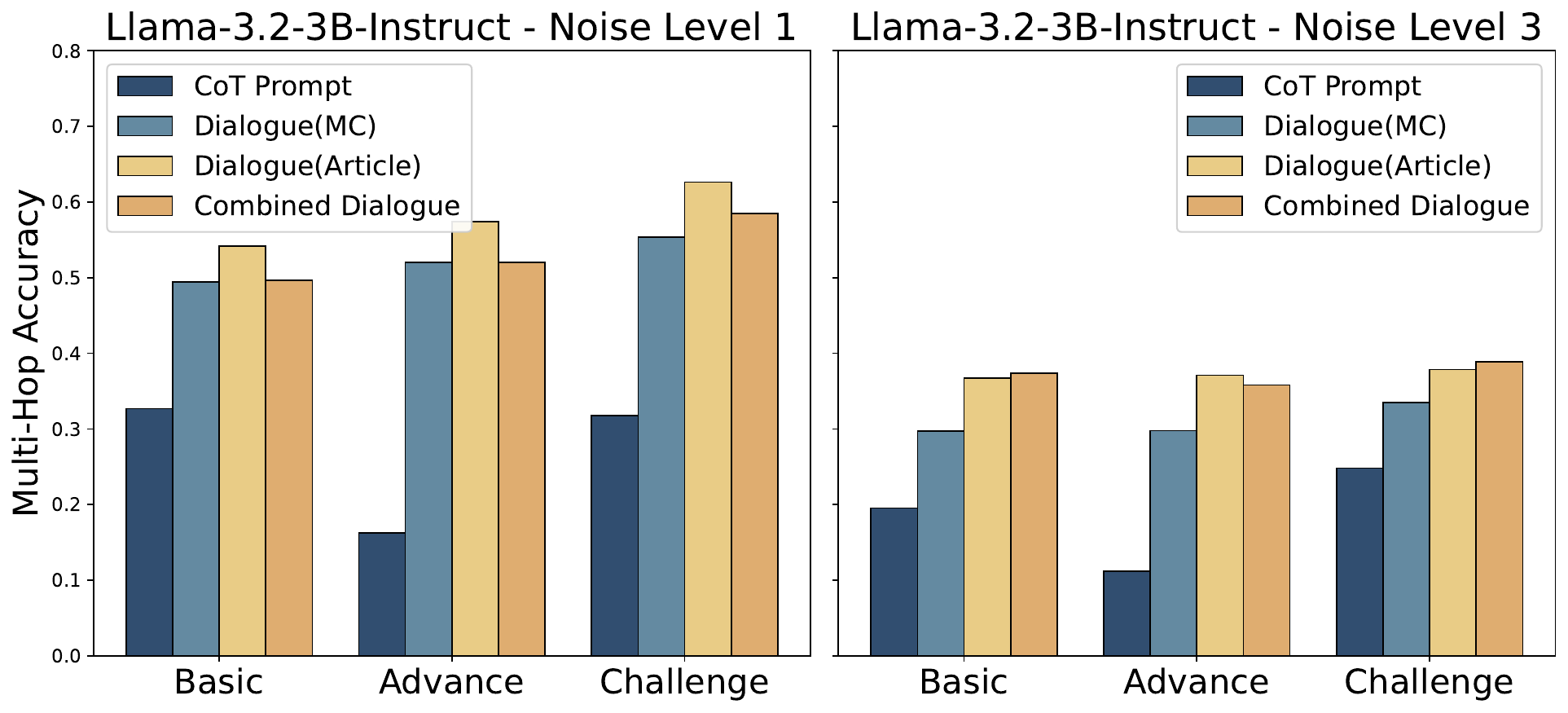} 
    \vspace{-6mm}
    \caption{\small Comparison between \texttt{dialogue-tuning} and Chain-of-Thought (CoT) prompting in the One-Round setting across Basic, Advanced, and Challenge tasks under Noise Levels 1 and 3.}
    \label{fig:cot}
    \vspace{-6mm}
\end{figure}

\textbf{\textit{Q1}: What is the effect of \texttt{dialogue-tuning} on reasoning across all difficulty levels in \textit{One-Round} setting? \textit{A1}: Improves reasoning across basic, advanced, and challenge levels.}

The results (\autoref{tab:basic}, \autoref{tab:advance}, \autoref{tab:challenge}) demonstrate that \texttt{dialogue-tuning} significantly enhances the reasoning performance of models compared to both multi-choice, article-based tuning and baseline strategies across varying levels of task difficulty (basic, advanced, and challenge). Compared to multi-choice tuning, \texttt{dialogue-tuning} shows significant improvements, with a 19.46\% higher Multi-Hop Accuracy and an 18.86\% increase in Single-Wise Accuracy at the basic level (see \autoref{tab:basic}). This advantage persists in other difficulty levels, where \texttt{dialogue-tuning} consistently outperforms multi-choice tuning. When tuning with case reports, \texttt{dialogue-tuning} does not exhibit as large of an improvement but still achieves a 2.54\% higher Multi-Hop Accuracy in the advanced setting and a 1.45\% increase in the challenge setting. Additionally, our baseline strategies perform notably worse than our combined \texttt{dialogue-tuning} approach. For example, in basic settings, the baseline achieves only 28.82\% Single-Wise Accuracy, whereas \texttt{dialogue-tuning} reaches 43.40\%. Even in the more challenging task, combined \texttt{dialogue-tuning} maintains a clear advantage, scoring 47.58\% (Single-Wise Accuracy) compared to the baseline's 39.23\%(Single-Wise Accuracy), and 30.02\% versus 22.04\% in another metric.

\begin{figure}[h!]
    \centering
    \includegraphics[scale=0.22]{figure/general.pdf} 
    \vspace{-3mm}
    \caption{\small Comparison of scores between baseline and combined dialogue approaches for LLaMA 3.2-3B Instruct and Qwen2.5-3B-Instruct across MedQA, MedMCQA, and PubMedQA datasets. The combined dialogue approach consistently improves performance.}
    \label{fig:general}
    \vspace{-3mm}
\end{figure}

\begin{table*}[h!]
\resizebox{\textwidth}{!}{%
\begin{tabular}{@{}ccccccccccc@{}}
\toprule
\multicolumn{11}{c}{Challenge (Multi-Hop Acc)} \\ \midrule
Model & \multicolumn{1}{c|}{Noise Level} & Raw & Multi-Choice & \multicolumn{1}{c|}{Dialogue(MC)} & Raw & Article & \multicolumn{1}{c|}{Dialogue(Article)} & Raw & Baseline & Combined Dialogue \\ \midrule
 & \multicolumn{1}{c|}{0} & 0.1019 & 0.0697 & \multicolumn{1}{c|}{\textbf{0.1154}} & 0.1019 & 0.1187 & \multicolumn{1}{c|}{\textbf{0.121}} & 0.1019 & 0.0888 & \textbf{0.1254} \\
 & \multicolumn{1}{c|}{1} & 0.2931 & 0.0698 & \multicolumn{1}{c|}{\textbf{0.4288}} & 0.2931 & 0.4449 & \multicolumn{1}{c|}{\textbf{0.5142}} & 0.2931 & 0.301 & \textbf{0.4725} \\
\multirow{-3}{*}{Llama-3.2-3B-Instruct} & \multicolumn{1}{c|}{3} & \textbf{0.227} & 0.0767 & \multicolumn{1}{c|}{0.2144} & 0.227 & 0.2577 & \multicolumn{1}{c|}{\textbf{0.267}} & 0.227 & 0.1935 & \textbf{0.2834} \\
 & \multicolumn{1}{c|}{0} & 0.1192 & 0.1031 & \multicolumn{1}{c|}{\textbf{0.1266}} & 0.1192 & \textbf{0.1232} & \multicolumn{1}{c|}{0.1217} & 0.1192 & 0.1087 & \textbf{0.125} \\
 & \multicolumn{1}{c|}{1} & 0.4037 & 0.347 & \multicolumn{1}{c|}{\textbf{0.4997}} & 0.4037 & \textbf{0.456} & \multicolumn{1}{c|}{0.4478} & 0.4037 & 0.3594 & \textbf{0.4935} \\
\multirow{-3}{*}{Qwen2.5-3B-Instruct} & \multicolumn{1}{c|}{3} & 0.2802 & 0.2062 & \multicolumn{1}{c|}{\textbf{0.2865}} & 0.2802 & 0.2769 & \multicolumn{1}{c|}{\textbf{0.2927}} & 0.2802 & 0.2712 & \textbf{0.3013} \\
\rowcolor[HTML]{CFF7F2} 
\multicolumn{2}{c|}{\cellcolor[HTML]{CFF7F2}Average} & 0.2375 & 0.1454 & \multicolumn{1}{c|}{\cellcolor[HTML]{CFF7F2}\textbf{0.2786}} & 0.2375 & 0.2796 & \multicolumn{1}{c|}{\cellcolor[HTML]{CFF7F2}\textbf{0.2941}} & 0.2375 & 0.2204 & \textbf{0.3002} \\ \midrule
\multicolumn{11}{c}{Challenge (Single-Wise Acc)} \\ \midrule
Model & \multicolumn{1}{c|}{Noise Level} & Raw & Multi-Choice & \multicolumn{1}{c|}{Dialogue(MC)} & Raw & Article & \multicolumn{1}{c|}{Dialogue(Article)} & Raw & Baseline & Combined Dialogue \\ \midrule
 & \multicolumn{1}{c|}{0} & 0.2026 & 0.118 & \multicolumn{1}{c|}{\textbf{0.2316}} & 0.2026 & 0.2314 & \multicolumn{1}{c|}{\textbf{0.2318}} & 0.2026 & 0.1588 & \textbf{0.2392} \\
 & \multicolumn{1}{c|}{1} & 0.5586 & 0.2013 & \multicolumn{1}{c|}{\textbf{0.6779}} & 0.5586 & 0.6848 & \multicolumn{1}{c|}{\textbf{0.7384}} & 0.5586 & 0.4599 & \textbf{0.6967} \\
\multirow{-3}{*}{Llama-3.2-3B-Instruct} & \multicolumn{1}{c|}{3} & 0.4357 & 0.1728 & \multicolumn{1}{c|}{\textbf{0.4549}} & 0.4357 & 0.4854 & \multicolumn{1}{c|}{\textbf{0.4898}} & 0.4357 & 0.3336 & \textbf{0.4945} \\
 & \multicolumn{1}{c|}{0} & 0.2115 & 0.1937 & \multicolumn{1}{c|}{\textbf{0.2217}} & 0.2115 & \textbf{0.2229} & \multicolumn{1}{c|}{0.2218} & 0.2115 & 0.1965 & \textbf{0.2154} \\
 & \multicolumn{1}{c|}{1} & 0.6758 & 0.6564 & \multicolumn{1}{c|}{\textbf{0.7356}} & 0.6758 & \textbf{0.7284} & \multicolumn{1}{c|}{0.7076} & 0.6758 & 0.7156 & \textbf{0.725} \\
\multirow{-3}{*}{Qwen2.5-3B-Instruct} & \multicolumn{1}{c|}{3} & 0.4629 & 0.4198 & \multicolumn{1}{c|}{\textbf{0.4824}} & 0.4629 & 0.4804 & \multicolumn{1}{c|}{\textbf{0.485}} & 0.4629 & \textbf{0.4895} & 0.4839 \\
\rowcolor[HTML]{CFF7F2} 
\multicolumn{2}{c|}{\cellcolor[HTML]{CFF7F2}Average} & 0.4245 & 0.2937 & \multicolumn{1}{c|}{\cellcolor[HTML]{CFF7F2}\textbf{0.4674}} & 0.4245 & 0.4722 & \multicolumn{1}{c|}{\cellcolor[HTML]{CFF7F2}\textbf{0.4791}} & 0.4245 & 0.3923 & \textbf{0.4758} \\ \bottomrule
\end{tabular}%
}
\caption{\small Performance of Llama-3.2-3B-Instruct and Qwen2.5-3B-Instruct under three tuning strategies across noise levels, evaluated on Challenge tasks in One-Round setting.}
\label{tab:challenge}
\end{table*}


\textbf{\textit{Q2}: Does \texttt{dialogue-tuning} still show the advantage in the multi-round setting?
\textit{A2}: Yes, it still outperforms traditional tuning methods.}

\begin{figure}[t]
  \centering

  \begin{subfigure}[b]{0.48\columnwidth}
    \includegraphics[width=\linewidth]{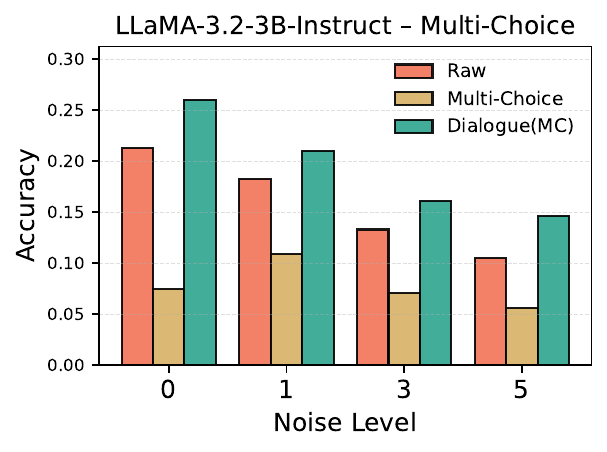}
  \end{subfigure}
  \hfill
  \begin{subfigure}[b]{0.48\columnwidth}
    \includegraphics[width=\linewidth]{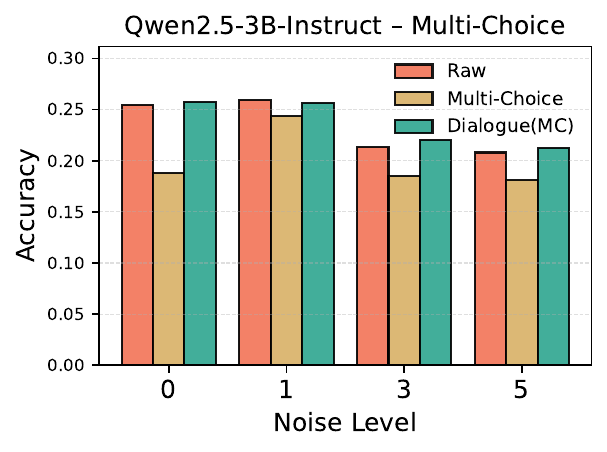}
  \end{subfigure}

  \begin{subfigure}[b]{0.48\columnwidth}
    \includegraphics[width=\linewidth]{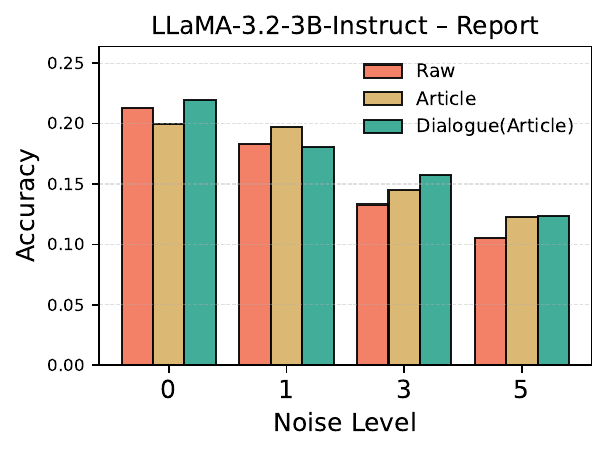}
  \end{subfigure}
  \hfill
  \begin{subfigure}[b]{0.48\columnwidth}
    \includegraphics[width=\linewidth]{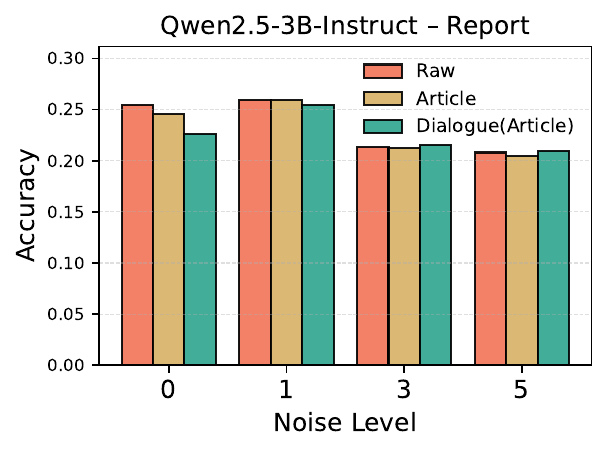}
  \end{subfigure}

  \begin{subfigure}[b]{0.48\columnwidth}
    \includegraphics[width=\linewidth]{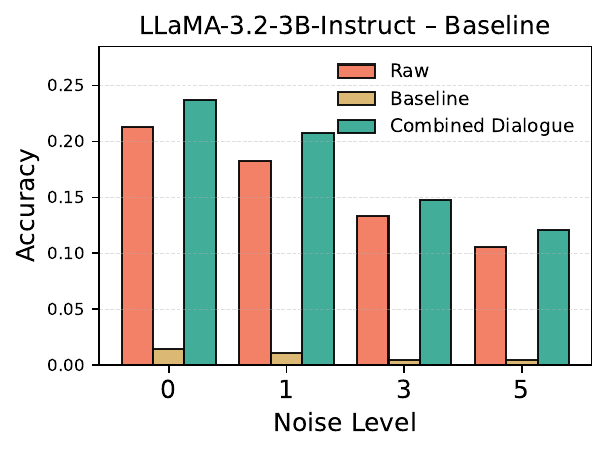}
  \end{subfigure}
  \hfill
  \begin{subfigure}[b]{0.48\columnwidth}
    \includegraphics[width=\linewidth]{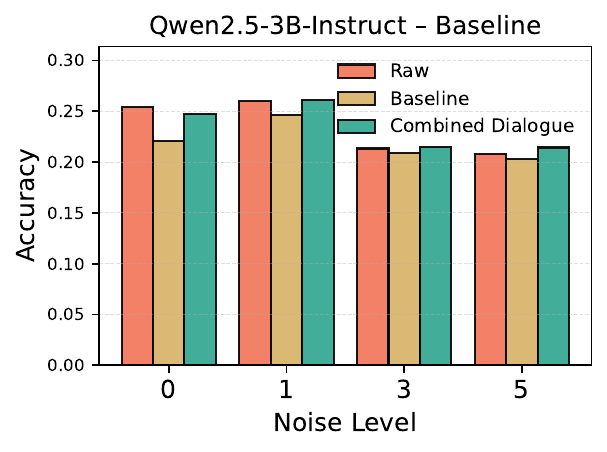}
  \end{subfigure}

  \caption{\small Performance of Llama-3.2-3B-Instruct and Qwen2.5-3B-Instruct under three tuning strategies across four noise levels.}
  \label{fig:cl_multi}
  \vspace{-2mm}
\end{figure}

In \autoref{fig:cl_multi}, we clearly demonstrate the performance of fine-tuned model in the multi-round setting across confusion levels 0 to 5. For the LLaMA 3.2-3B Instruct, our \texttt{dialogue-tuning} exhibits clear advantages across all three experimental settings, and further enhances performance in multi-round tasks—achieving a 4.06\% improvement even in the noisiest environment (level 5). As for the Qwen 2.5-3B Instruct model, while it does not show as significant a gap compared to LLaMA, it still maintains an advantage, particularly in high-noise environments. Specifically, it achieves a 3.16\% improvement over multi-choice tuning strategies at confusion level 5.

\begin{figure}[h!]
    \centering
    \includegraphics[scale=0.24]{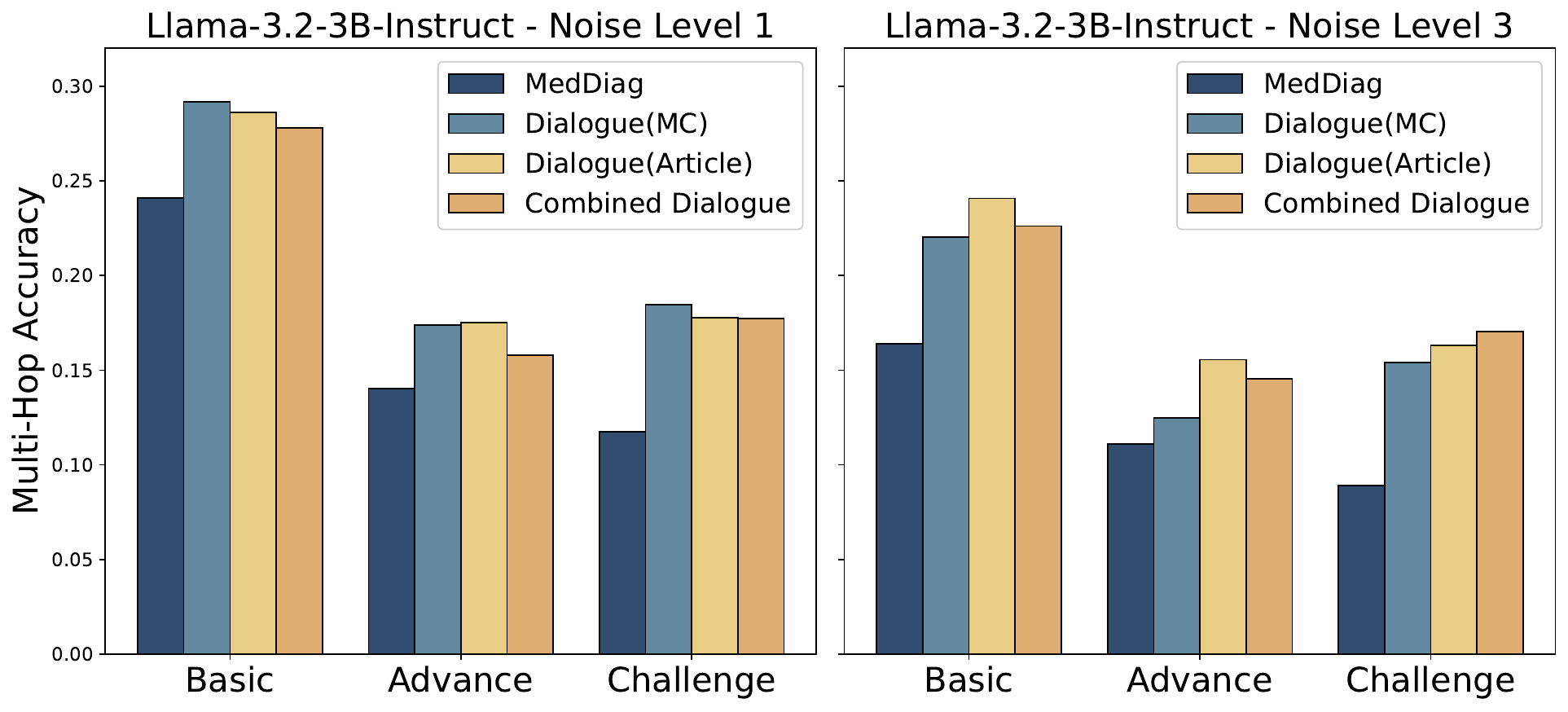} 
    \vspace{-4mm}
    \caption{\small Comparison between \texttt{dialogue-tuning} and MedDiag-tuned models in the Multi-Round setting across three difficulty levels under Noise Levels 1 and 3.}
    \label{fig:meddiag}
    \vspace{-3mm}
\end{figure}

\textbf{\textit{Q3}: How does \texttt{dialogue-tuning}compare to existing reasoning and dialogue-based training approaches?
\textit{A3}: It achieves more robust reasoning than CoT and MedDiag baselines while generalizing well to standard QA tasks.}

As shown in Figure~\ref{fig:cot}, dialogue-tuned models consistently surpass CoT prompting across all difficulty levels and noise conditions, indicating that reasoning-oriented dialogue supervision provides stronger inductive bias than explicit step-by-step prompting. In the Multi-Round setting (Figure~\ref{fig:meddiag}), our synthetic dialogue-tuned models also match or exceed the performance of models trained on MedDiag ~\cite{zeng-etal-2020-meddialog} conversations, suggesting that strategic dialogue reformulation effectively captures diagnostic reasoning structure. Finally, as shown in Figure~\ref{fig:general}, dialogue-tuned models generalize well to widely used general medical QA datasets (MedQA~\cite{yao2024medqa}, MedMCQA~\cite{pal2022medmcqa}, PubMedQA~\cite{jin2019pubmedqa}), confirming the robustness and transferability of our approach.


\begin{figure}[h!]
    \centering
    \includegraphics[scale=0.24]{figure/human_eval.pdf} 
    \vspace{-3mm}
    \caption{\small Human Evaluation Performance.}
    \label{fig:human}
    \vspace{-7mm}
\end{figure}

\subsection{Human Evaluation}
\label{sec:human_eval}
To assess the fidelity of our LLM-generated dialogues to the original context, we conducted a human evaluation study with 32 participants in U.S., including medical students. Participants rated each dialogue on a 5-point scale ranging from Fully Covered (4: all essential information preserved) to Not Covered (0: core information missing or distorted).

The results demonstrated strong performance, shown in \autoref{fig:human}: 79\% of the MC dialogues and 74\% of the Report dialogues were rated as Fully Covered (score 4), indicating that the generated dialogues preserved all key information from the original context. Notably, only 3–4\% of outputs fell into the Minimally/Not Covered categories (scores 0–1), suggesting rare failures in coherence. These findings confirm that our LLM-generated dialogues are highly faithful to the source material, achieving the primary goal of dependency on and comprehensive coverage of the raw context. The human evaluation thus validates the reliability of our approach for producing trustworthy dialogue outputs.
\vspace{-5mm}
\section{Conclusion}
In this work, we introduced \muddymaze, a novel benchmark for evaluating the logical reasoning and evidence-based decision-making abilities of medical language models under realistic clinical uncertainty. By reframing medical QA as an evidence-ranking problem, \muddymaze captures both holistic and stepwise reasoning behaviors that are critical in diagnostic practice. We further proposed \texttt{dialogue-tuning}, a training strategy that reformulates static QA and article data into simulated doctor–patient interactions, where the model is updated only on the physician’s turns. Comprehensive experiments demonstrate that our method is both effective and efficient.

This work highlights the importance of a dynamical approach to advancing reasoning in medical AI systems. \texttt{Dialogue-tuning} aligns training with the step-by-step cognitive processes required for diagnostic decision-making, providing a framework for developing more reliable models.

\section*{Limitations}
The dialogue generation process, which relies solely on Gemini-2.5 Flash, may introduce certain biases. Even though we randomly sample some of the generated dialogues for human evaluation, relying on a single large language model for dialogue generation could lead to model-specific biases—particularly in how it structures conversations and prioritizes certain types of medical information. We appreciate this important point. To encourage diversity and robustness, our benchmark incorporates samples from multiple sources (MedQA, MedBullets, JAMA Challenge), each with varied disease coverage and linguistic styles.

Additionally, the noise injection and multi-step reasoning design of \muddymaze introduce substantial unpredictability and complexity, mirroring unconventional and unpredictable situations that appear in real-world clinical decisions.

We acknowledge that more explicit evaluation of rare diseases or population shifts would strengthen our claims. As part of future work, we plan to extend our framework to clinical case simulations derived from local hospital EHR data, enabling evaluation on broader demographics and out-of-distribution cases.

\section*{Acknowledgement}
Dr. Qingyu Chen is supported by the National Institutes of Health National Library of Medicine under Award Number R01LM014604.
\bibliography{custom}

\clearpage
\appendix

\section{Dialogue Generation}
\label{sec:appA}
\subsection{Why We Need Dialogue Tuning?}
\texttt{Dialogue-tuning} is proposed as a more effective approach for capturing logical relationships, as the dialogue format inherently mirrors the reasoning process found in human doctor-patient interactions.

\textbf{Interactive Nature of Dialogue.} The dialogue format enables iterative, question-and-answer reasoning that mimics real diagnostic processes. Clinicians progressively gather information through targeted questioning, with each response reducing diagnostic uncertainty. This stepwise approach organizes clinical data logically, making the reasoning chain more transparent and tractable for AI systems.

\textbf{Empirical Support for Dialogue Tuning.} Our experiments in \autoref{sec:experiment} demonstrate that \texttt{dialogue-tuning} significantly outperforms multi-choice and article-based tuning in tasks designed with our benchmark. Key findings include:
\ding{182} Dialogue-tuned models exhibit higher accuracy in reasoning tasks, including those requiring multi-step integration of evidence.
\ding{183} Dialogue-tuned models remain robust even when tackling challenging tasks or integrating noisy inputs, maintaining performance where other approaches falter.
\ding{184} They also have competitive capabilities in general medical question-answering tasks, as the model is fine-tuned with multiple-choice QA and medical articles.

\subsection{Prompt Templates}

\begin{tcolorbox}[breakable]
\small
\ttfamily
\textbf{Multi-choice QA to Dialogue:} Please convert the following paragraph into a doctor-patient dialogue. 
Ensure that all the information provided, including personal details, symptoms, 
examination findings, diagnosis, and treatment, is included. Most important is 
the final answer, "\{\{Answer\}\}", which must be included in the dialogue 
without any changes. Use natural conversational language to connect the 
details, but do not introduce any new information. 
The dialogue should not be too redundant:

\{\{Document\}\}
\end{tcolorbox}

\begin{tcolorbox}[breakable]
\small
\ttfamily
\textbf{Article to Dialogue:}
Please convert the following paragraph into a doctor-patient dialogue. 
Ensure that all the information provided, like personal details, symptoms, 
examination findings, diagnosis, and treatment, is included. 
Use natural conversational language to connect the details, 
but do not introduce any new information. 
The dialogue should not be too redundant:

\{\{Document\}\}
\end{tcolorbox}

\subsection{Task Reformulation Examples}
This example demonstrates how we use the above prompt to guide Gemini 2.5-flash in transitioning from multiple-choice question answering to a doctor-patient dialogue format.
\begin{tcolorbox}[breakable]
\small
\ttfamily
A 23-year-old pregnant woman at 22 weeks gestation presents with burning upon urination. She states it started 1 day ago and has been worsening despite drinking more water and taking cranberry extract. She otherwise feels well and is followed by a doctor for her pregnancy. Her temperature is 97.7°F (36.5°C), blood pressure is 122/77 mmHg, pulse is 80/min, respirations are 19/min, and oxygen saturation is 98\% on room air. Physical exam is notable for an absence of costovertebral angle tenderness and a gravid uterus. Which of the following is the best treatment for this patient?
\end{tcolorbox}
\begin{tcolorbox}[breakable]
\small
\ttfamily
Doctor: Hi there, I understand you're having some trouble. What brings you in today?\\
Patient: Hi Doctor. I'm 22 weeks pregnant, and for the past day, I've had burning when I pee. It's really uncomfortable.\\
Doctor: I see. And how long has this been going on?\\
Patient: Just since yesterday, and it's actually getting worse, even though I've been drinking a lot of water and taking cranberry extract.\\
Doctor: I understand. Other than the burning, how are you feeling?\\
Patient: I feel fine otherwise. My pregnancy is being followed regularly by my OB.\\
Doctor: Okay, let's take a look. I'm going to check your vitals and do a quick exam. Your temperature is 97.7, blood pressure is 122 over 77, pulse is 80, respirations are 19, and your oxygen is 98\%. Everything looks good there. I'm also checking for any tenderness in your back near your kidneys... that's good, no tenderness. And I can feel your uterus is growing as expected. Given your symptoms and the examination, the best treatment for this is Nitrofurantoin.
\end{tcolorbox}

\subsection{Tuning Details}
In our tuning strategies, we consistently use LoRA tuning with the following parameters: rank = 8, lora\_alpha = 32, and lora\_dropout = 0.1. In all three comparison studies, both the control group and the experimental group share the same hyperparameters. In the \textit{Multi-Choice Comparison Study}, we set Qwen2.5-3B to 2 epochs, a learning rate of $5\times10^{-4}$, and a max length of 1024; LLaMA-3.2-3B to 2 epochs, a learning rate of $5\times10^{-4}$, and a max length of 1024. In the \textit{Article Comparison Study}, we configure Qwen2.5-3B with 2 epochs, a learning rate of $1\times10^{-5}$, and a max length of 1024; LLaMA-3.2-3B with 2 epochs, a learning rate of $5\times10^{-5}$, and a max length of 2048. Finally, in the \textit{Combined Dialogue and Baseline Study}, we assign Qwen2.5-3B to 2 epochs, a learning rate of $2\times10^{-5}$, and a max length of 2048; LLaMA-3.2-3B to 2 epochs, a learning rate of $5\times10^{-5}$, and a max length of 2048.

\textbf{Testbed.}
We fine-tuned the Llama 3.2-3B Instruct and Qwen 2.5-3B Instruct models using 2 NVIDIA RTX 6000 GPUs, each with 48GB of memory. 
We running our benchmark also on the NVIDIA RTX 6000 48GB GPUs.

\section{Benchmark: \muddymaze}
\label{sec:appendix}
\textbf{Dataset.} Our benchmark integrates the MedQA-USMLE Test Set, Medbullets, and JAMA Clinical Challenge. The fine-tuning is based on the MedQA-USMLE Train Set, which have around 10.2k question-answer pairs and approximately 12k PubMed articles. All dataset information shows in \autoref{fig:datainfo}.

\subsection{Clinical and Examination Basis for \muddymaze Benchmark}
The design of \muddymaze is rigorously aligned with established medical licensing exams and real-world diagnostic workflows. Below, we outline its foundations in several key areas:

\noindent $\bullet$ \textbf{USMLE Step 2 Clinical Skills (CS).} It required examinees would take a history, perform a physical exam, formulate differential diagnoses, and write a patient note. 

\noindent $\bullet$ \textbf{USMLE Step 3 CCS.} It assesses clinical decision-making through Computer-based Case Simulations (CCS). These simulations require doctors to diagnose and manage patients by sequentially ordering tests, interpreting results, and initiating treatments—all while filtering out irrelevant information (like incidental findings or patient anecdotes) that could distract from critical decisions.

\noindent $\bullet$ \textbf{Medical Jeopardy competitions.} An answer-first format, where contestants hear a clinical "clue" (e.g., "This tumor causes episodic hypertension and headaches") and must respond with the correct question (e.g., "What is pheochromocytoma?"). It required clinicians compete to solve clinical puzzles by connecting fragmented clues—such as symptoms, labs, or imaging findings—into accurate diagnoses. Contestants must rapidly prioritize key evidence while ignoring distractors, mirroring real-world diagnostic reasoning.

The design of \muddymaze integrates core principles from these real-world clinical assessments:

$\bullet$ USMLE Step 2 CS’s iterative data gathering (history → exam → tests) inspired our multi-round evidence ranking, where models must simulate a clinician’s stepwise reasoning.

$\bullet$ USMLE Step 3 CCS’s emphasis on prioritizing actions amid distractions (e.g., ignoring incidental findings) directly aligns with \muddymaze’s noise injection and dynamic evidence selection.

$\bullet$ Medical Jeopardy’s answer-first format—requiring contestants to reverse-engineer diagnoses from clues—parallels our one-round evidence chaining, where models reconstruct logical sequences (e.g., lab → imaging → diagnosis) from fragmented inputs.

Together, these connections validate \muddymaze’s clinical fidelity, ensuring it tests not just medical knowledge, but the decision-making workflows and noise resilience essential in practice.
\subsection{Prompt Template}
\begin{tcolorbox}[breakable]
\small
\ttfamily
Here is the background information: "\{\{prerequisit\}\}"\\
Question: \{\{question\}\}\\
Answer: \{\{answer\}\}\\
Below are several evidence sentences. Identify the \{\{groundtruth zoo length\}\} sentences that, if added to the background information, would support inferring the answer based on the given question-answer pair. Please choose the sentence in logical order!\\
\{\{tagged maze\}\}\\
Provide only the indices of the relevant sentences in brackets formatted like this: [ ], no more than \{\{groundtruth zoo length\}\} sentences.\\
ANSWER:
\end{tcolorbox}

\begin{tcolorbox}[breakable]
\small
\ttfamily
Here is the background information: "\{\{prerequisit\}\}"\\
Question: \{\{question\}\}\\
Answer: \{\{answer\}\}\\
Below are several evidence sentences. Based on the given question-answer pair, please select which sentence should be added to the background information to support inference of the answer. \\
\{\{tagged maze\}\}\\
You have \{\{groundtruth zoo length\}\} attempts in total to make a selection; this is your \{\{i\_th\}\} attempt. Please choose the sentence in logical order!\\
Provide only the indices of the relevant sentences in brackets formatted like this: [ ]\\
ANSWER:
\end{tcolorbox}
\subsection{Task Reformulation Examples}
This is an example of transitioning from the traditional question-answering task to our benchmark. The results are generated using LLaMA 3.1-8B at the basic task level, with a noise level of 0, in a one-round setting.
\begin{tcolorbox}[breakable]
\small
\ttfamily
A 67-year-old man with transitional cell carcinoma of the bladder comes to the physician because of a 2-day history of ringing sensation in his ear. He received this first course of neoadjuvant chemotherapy 1 week ago. Pure tone audiometry shows a sensorineural hearing loss of 45 dB.\\
Question: The expected beneficial effect of the drug that caused this patient's symptoms is most likely due to which of the following actions?\\
Answer: Cross-linking of DNA
\end{tcolorbox}
\begin{tcolorbox}[breakable]
\small
\ttfamily
Here is the background information: \"A 67-year-old man with transitional cell carcinoma of the bladder comes to the physician because of a 2-day history of ringing sensation in his ear.\"\\
Question: The expected beneficial effect of the drug that caused this patient's symptoms is most likely due to which of the following actions?\\
Answer: Cross-linking of DNA\\
Below are several evidence sentences. Identify the 2 sentences that, if added to the background information, would support inferring the answer based on the given question-answer pair. Please choose the sentence in logical order!\\
0: Pure tone audiometry shows a sensorineural hearing loss of 45 dB.\\
1: He received this first course of neoadjuvant chemotherapy 1 week ago.\\
Provide only the indices of the relevant sentences in brackets formatted like this: [ ], no more than 2 sentences.\\
ANSWER:
[1], [0]
\end{tcolorbox}

\section{Error Analysis}
\label{app:error}
\subsection{Case Analysis: Comparison of RAW, CoT-Prompt, and Dialogue-Tuned Models}

\begin{enumerate}
    \item \textbf{Raw Model:} The raw model often produces shallow, pattern-based outputs that loosely connect keywords like “chemotherapy” and “hearing loss” without deep reasoning. \textbf{Incorrect}
    
    \item \textbf{COT-prompt Model:} Despite generating a coherent verbal explanation under the “Let’s think step by step” prompt, the CoT model fails to select the correct evidence order. It narrates a cause $\rightarrow$ effect chain but ultimately chooses \texttt{[0, 1]}, reversing that logic. This illustrates a common CoT failure: the explanation and the action (evidence selection) are decoupled. The model focuses more on sounding logical than being structurally accurate. \textbf{Incorrect}
    
    \item \textbf{Dialogue Tuning Model:} In contrast, the dialogue-tuned model demonstrates structured, clinically grounded reasoning: it identifies the treatment (chemotherapy) as the initiating factor, links it to a known adverse effect (sensorineural hearing loss), and leverages the timing of symptom onset to strengthen the causal inference. The selected evidence—[1], [0]—is minimal, correctly ordered, and directly supports the answer. \textbf{Correct}
\end{enumerate}
\vspace{-3mm}
\paragraph{Conclusion}
In summary, while raw and CoT-prompted models may produce plausible answers or fluent explanations, they often lack structured, causal reasoning. Dialogue-tuned models, by contrast, learn to reason in a clinically faithful, stepwise manner—aligning explanation, evidence selection, and causal order. This leads to more accurate, interpretable, and reliable multi-step inference in medical tasks.

\subsection{Supplementary Outputs}
\vspace{3mm}

\begin{tcolorbox}[fontupper=\small,title=Question and GroundTruth]
Here is the background information: "A 67-year-old man with transitional cell carcinoma of the bladder comes to the physician because of a 2-day history of ringing sensation in his ear." Question: The expected beneficial effect of the drug that caused this patient's symptoms is most likely due to which of the following actions? Answer: Cross-linking of DNA\\\
Below are several evidence sentences. Identify the 2 sentences that, if added to the background information, would support inferring the answer based on the given question-answer pair. Please choose the sentence in logical order! 0: Pure tone audiometry shows a sensorineural hearing loss of 45 dB. 1: He received this first course of neoadjuvant chemotherapy 1 week ago. Answer: [1,0]
\end{tcolorbox}

\begin{tcolorbox}[fontupper=\small,title=Raw Model Output]
[0], [1]\\\
These sentences support inferring that the drug caused the patient's symptoms because they provide evidence of a sensorineural hearing loss, which is consistent with ototoxicity, a known side effect of certain chemotherapy agents.
\end{tcolorbox}

\begin{tcolorbox}[fontupper=\small,title=COT-prompt Model Output]
To answer this question, let's analyze the situation step by step:\\\\
1. The patient has transitional cell carcinoma of the bladder and is experiencing a ringing sensation in his ear (tinnitus), which is a common side effect of certain chemotherapy drugs.\\
2. The patient recently received a course of neoadjuvant chemotherapy, which suggests that the tinnitus is likely related to this treatment.\\
3. The question asks about the expected beneficial effect of the drug that caused the patient's symptoms, which is tinnitus.\\
4. The answer is ``Cross-linking of DNA'', which is a mechanism of action associated with certain chemotherapy drugs, such as platinum-based agents (e.g., cisplatin).\\\\
Now, let's examine the evidence sentences:\\\\
0: Pure tone audiometry shows a sensorineural hearing loss of 45 dB.\\
This sentence suggests that the patient's hearing loss is related to the chemotherapy, but it doesn't directly support the answer.\\\\
1: He received this first course of neoadjuvant chemotherapy 1 week ago.\\
This sentence confirms that the patient recently received chemotherapy, which is consistent with the patient's symptoms.\\\\
To support inferring the answer, we need another sentence that connects the chemotherapy to the mechanism of action mentioned in the answer. Sentence 0 provides this connection by mentioning the hearing loss, which is a known side effect of certain chemotherapy drugs that cause cross-linking of DNA.\\\\
The relevant sentences are: \\
{[0], [1]}
\end{tcolorbox}

\begin{tcolorbox}[fontupper=\small,title=Dialogue-Tuning Model Output]
[1], [0]\\\
These sentences support the inference that the drug causing the patient's symptoms is likely cisplatin, a common chemotherapeutic agent used to treat transitional cell carcinoma of the bladder. Cisplatin is known to cause ototoxicity, leading to sensorineural hearing loss, and its ototoxic effects can be detected through pure tone audiometry. The timing of the chemotherapy (1 week ago) also supports this inference, as cisplatin's ototoxic effects typically manifest within a week of treatment.
\end{tcolorbox}

\subsection{Error Analysis}
Here is an example:
\begin{tcolorbox}[breakable]
\small
\ttfamily
Here is the background information: ``A 46-year-old man is brought to the emergency department for evaluation of altered mental status.'' Question: Which of the following is the most likely cause of this patient's symptoms? Answer: Benzodiazepine intoxication. 

Below are several evidence sentences. Identify the 8 sentences that, if added to the background information, would support inferring the answer based on the given question-answer pair. Please choose the sentence in logical order!

0: On physical examination, an alcoholic smell and slurred speech are noted. \\
1: An ECG shows no abnormalities. \\
2: His pupils are normal. \\
3: He is somnolent but responsive when aroused. \\
4: Neurological exam shows diminished deep tendon reflexes bilaterally and an ataxic gait. \\
5: Blood alcohol concentration is 0.04\%. \\
6: He was found on the floor in front of his apartment. \\
7: His pulse is 64/min, respiratory rate is 15/min, and blood pressure is 120/75 mm Hg. 

Then provide only the indices of the relevant sentences in brackets formatted like this: [ ], no more than 8 sentences. ANSWER:
\end{tcolorbox}

\begin{itemize}
    \item \textbf{Model Answer:} [0, 3, 4, 7, 6, 5, 1, 2]
    \item \textbf{Ground Truth:} [6, 3, 7, 0, 4, 2, 5, 1]
\end{itemize}

In this example, although our model is trained to capture clinical reasoning patterns, the nature of the evidence set poses a particular challenge. Many of the supporting sentences (e.g., 2, 5, 1) are clinically relevant but less explicit in their reasoning cues, requiring subtle inferences rather than obvious trigger words. This makes it harder for a smaller 3B-parameter model to consistently identify and order such evidence correctly. The model performs well on more straightforward cases but may struggle when clinical logic is embedded in finer-grained or seemingly trivial details. Enhancing the model’s ability to handle such nuanced clinical reasoning remains a key direction for our future work.

\section{Details on Human evaluation and Human Annotation}
Here is annotator agreement and survey introduction:
\begin{tcolorbox}[breakable]
\small
\ttfamily
Welcome and thank you for participating! In this questionnaire, we’re evaluating how well our system, which uses a LLM, transforms original paragraphs into dialogue format. Your feedback will help us assess the quality of the generated dialogues and improve future outputs. Please rate each dialogue based on how well it preserves the essential information from the original paragraph, using the following scale: 4 – Fully Covered: All essential information is present and clearly conveyed. 3 – Mostly Covered: Most key ideas are included, with minor omissions. 2 – Moderately Covered: About half of the vital information is included. 1 – Minimally Covered: Only a few important points are mentioned. 0 – Not Covered: The dialogue misses or distorts the core information. A score of 4 means the dialogue is excellent; 0 means it’s far off the mark. At the end of the questionnaire, there will be space for you to leave any suggestions or thoughts on how we can improve the prompts or the dialogue generation process. We truly appreciate your time and insights! If you have any questions, feel free to reach out to us anytime.
\end{tcolorbox}
One error case analysis is shown below:
\textit{Doctor: “Unfortunately, you passed away three months after the onset of the right-shoulder swelling.}

Here the patient is simultaneously alive (able to converse) and deceased, revealing a temporal-logic breakdown in the LLM’s generation. Such errors slipped through the coverage rubric because all factual points from the source paragraph were indeed “present,” yet their combination was nonsensical.

Moreover, we conducted an additional annotation on a subset of 500 dialogues with the help of 20 domain experts across five specialties—Infectious Disease, Pediatrics, Oncology, Ophthalmology, and Neurology. Our previous evaluation covered 300 samples. The results reflect average agreement among annotators rather than single-annotator judgments, ensuring high reliability. In this larger set, 76\% of sampled dialogues were rated as “fully covered” with clinically coherent reasoning.

\end{document}